\pdfoutput=1
\documentclass[11pt]{article}

\usepackage[]{ACL2023}
\usepackage{subcaption}

\usepackage{times}
\usepackage{latexsym}

\usepackage[T1]{fontenc}
\usepackage[utf8]{inputenc}

\usepackage{graphicx,xcolor}  % グラフィックス関連
\usepackage{pxrubrica}        % ルビ
\usepackage{inconsolata}
\usepackage{url}
\usepackage{linguex}
\usepackage{ascmac}
\usepackage{booktabs}
\usepackage{dcolumn}

\title{MultiTool-CoT: GPT-3 Can Use Multiple External Tools with Chain of Thought Prompting}

\author{
Tatsuro Inaba$^1$ \hspace{1em}
Hirokazu Kiyomaru$^1$ \hspace{1em}
Fei Cheng$^1$ \hspace{1em} 
Sadao Kurohashi$^{1,2}$\\
$^1$Kyoto University, Japan \hspace{1em}
\\
$^2$National Institute of Informatics, Japan \hspace{1em}
\\
\texttt{\{inaba, kiyomaru, feicheng, kuro\}@nlp.ist.i.kyoto-u.ac.jp}\\
}

\begin{document}
\maketitle
\begin{abstract}

Large language models (LLMs) have achieved impressive performance on various reasoning tasks.
% To further improve the performance, we propose an interactive framework for using multiple external tools, such as a calculator and a knowledge retriever, during the reasoning process.
To further improve the performance, we propose MultiTool-CoT, a novel framework that leverages chain-of-thought (CoT) prompting to incorporate multiple external tools, such as a calculator and a knowledge retriever, during the reasoning process.
We apply MultiTool-CoT to the Task 2 dataset of NumGLUE, which requires both numerical reasoning and domain-specific knowledge.
The experiments show that our method significantly outperforms strong baselines and achieves state-of-the-art performance.
\footnote{Our code is publicly available at \url{https://github.com/InabaTatsuro/MultiTool-CoT}.}

\end{abstract}

\section{Introduction}

Reasoning refers to the logical process of inferring unknown facts from known facts.
Solving reasoning problems requires language understanding, real-world knowledge, arithmetic calculation, and symbolic processing.
Improving the reasoning capability of artificial intelligence has been a long-standing challenge and remains an active research topic to this day~\cite{gordon-etal-2012-semeval,sap-etal-2020-commonsense}.

Recently, large language models (LLMs) have achieved amazing performance on various reasoning tasks~\cite{GPT-3,lewkowycz2022solving,opt,palm}.
However, the amount of real-world knowledge learned by LLMs is still constrained by the size of model parameters and the training data.
This problem could be more severe in the case of sparse domain-specific knowledge.
Furthermore, LLMs are based on the computation among continuous token representations, which cannot ensure accurate arithmetic calculations.

To solve these problems, previous studies propose to complement the capabilities of LLMs with an external tool, such as a web browser or a calculator~\cite{2021-nakano-webgpt,dentaku,ReAct}.
This is performed by invoking an external tool during reasoning with LLMs and injecting the results into the reasoning process.
However, previous studies have focused on using a single external tool to solve a single problem with LLMs and have not addressed different problems together.

\begin{figure*}[t]
    \centering
    \includegraphics[width=\textwidth]{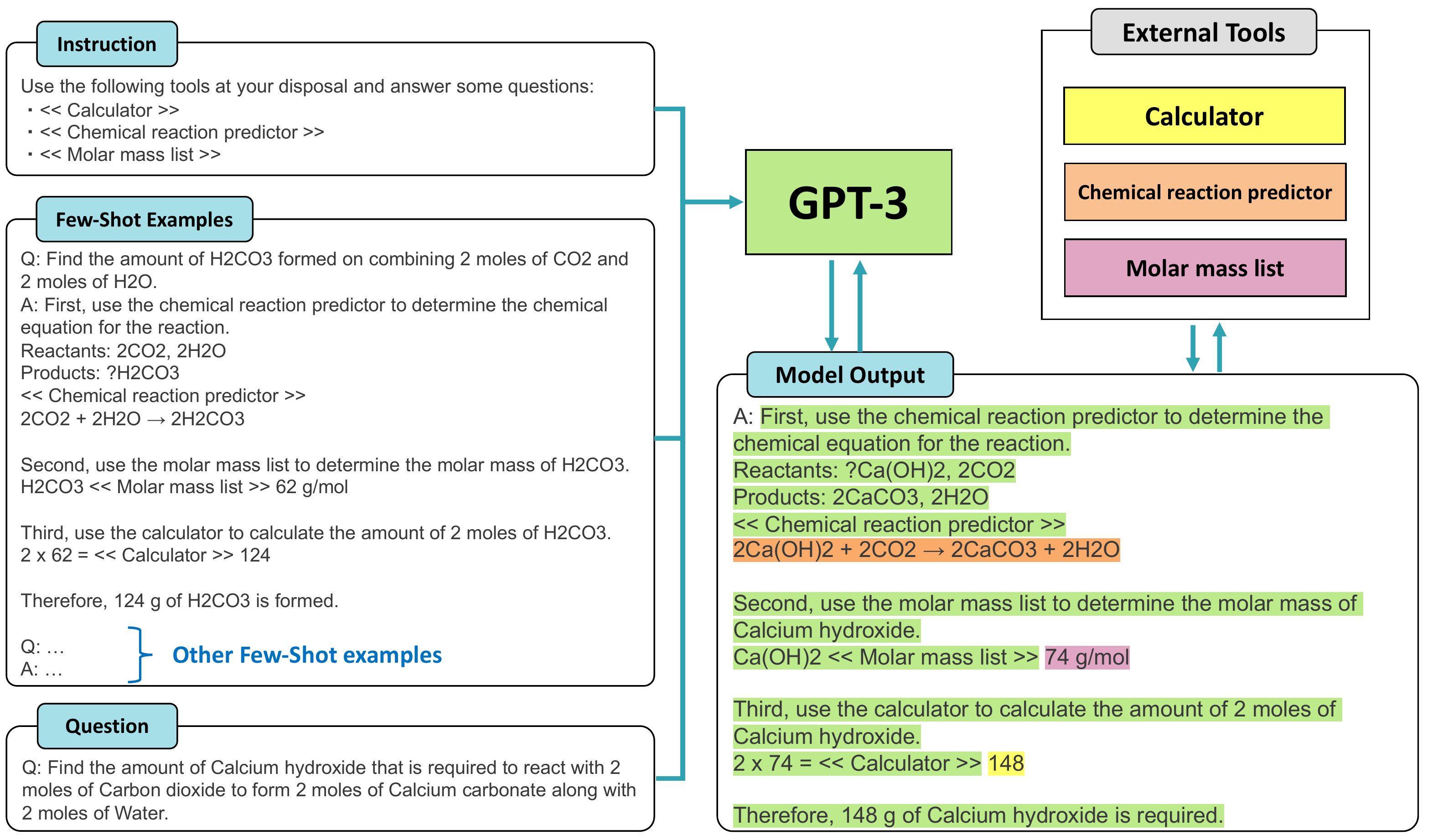}
    \caption{
        Overview of the MultiTool-CoT.
        The output of GPT-3, the calculator, the chemical reaction predictor, and the molar mass list are highlighted in green, yellow, orange, and purple, respectively.
    }
    \label{fig:overview}
\end{figure*}

This paper proposes MultiTool-CoT, an interactive framework that allows LLMs to use multiple external tools during reasoning.
Figure~\ref{fig:overview} provides an overview.
% In our framework, LLMs solve reasoning problems by generating reasoning processes including stop indicators to invoke external tools.
In MultiTool-CoT, LLMs solve reasoning problems by generating reasoning processes including tool triggers to invoke external tools.
We let LLMs learn to invoke multiple external tools at proper reasoning steps by chain-of-thought (CoT) prompting based on few-shot learning~\cite{wei2022chain}.

% As a proof of concept, we apply our framework to the Task 2 dataset of NumGLUE~\cite{mishra-etal-2022-numglue}, which requires both arithmetic reasoning and domain-specific knowledge.
As a proof of concept, we apply MultiTool-CoT to the Task 2 dataset of NumGLUE~\cite{mishra-etal-2022-numglue}, which requires both numerical reasoning and domain-specific knowledge.
Experiments show that MultiTool-CoT significantly outperforms strong baselines and achieves state-of-the-art performance.

\section{Related Work}
Large language models (LLMs) can perform various tasks by \textit{prompting}~\cite{10.1145/3560815}.
As for reasoning tasks, chain-of-thought (CoT) prompting~\cite{wei2022chain,kojima2022large} is known for its effectiveness, which elicits the answer with intermediate reasoning steps from LLMs.

There is a growing body of work on using an external tool to improve reasoning with LLMs.
\citet{dentaku} use a calculator to process mathematical formulas that appear in reasoning processes by fine-tuning LLMs to generate mathematical formulas with a tool trigger to call the calculator.
\citet{2021-nakano-webgpt} allow LLMs to use a web browser by fine-tuning LLMs to generate action codes to operate the browser.
Previous studies focus on a single problem of LLMs, namely, error-prone arithmetic calculation or incomplete real-world knowledge, and address it by fine-tuning LLMs so that they can call a single external tool.
In contrast, this study addresses multiple problems together by allowing LLMs to use multiple external tools.
Besides, this study presents a few-shot learning-based framework~\cite{GPT-3} for doing this, which does not require fine-tuning.

A very recent study~\cite{ReAct} proposes a few-shot learning-based method for invoking a Wikipedia API to perform knowledge-intensive reasoning tasks.
However, this study has not investigated the effectiveness of using multiple external tools.
A Python library named LangChain\footnote{\url{https://langchain.readthedocs.io/en/latest}} implements a framework for allowing LLMs to use multiple external tools based on \citet{ReAct}, which is similar to ours.
However, its effectiveness has not been investigated in any benchmark datasets as of this submission.

\section{Method}
\label{sec:proposed}

We propose MultiTool-CoT, an interactive framework that allows LLMs to use multiple external tools during reasoning.
Figure~\ref{fig:overview} illustrates an overview.

MultiTool-CoT leverages chain-of-thought (CoT) prompting based on few-shot learning~\cite{wei2022chain}.
Our prompt consists of an instruction specifying the available external tools, few-shot examples demonstrating several question-answer pairs with reasoning processes, and a question to be solved.
We manually annotate the reasoning processes shown as few-shot examples with tool triggers marked with corresponding input data, adhering to a specific format.
In this study, we let the string \texttt{$<<$External tool name$>>$} be a tool trigger.
For example, if we use a calculator as an external tool, we annotate the reasoning processes with the tool trigger \texttt{$<<$Calculator$>>$} after input formulas like $2 \times 62$.

When reasoning, GPT-3 follows the prompt and generates a reasoning process including tool triggers.
If a tool trigger is generated, we stop text generation.
We then extract the name of the external tool and the input for the tool from the reasoning process, execute the tool with the input, and append the result to the end of the reasoning process.
After that, we restart text generation.

If we cannot execute an external tool for some reason (e.g., invalid tool input is generated), we fall back on GPT-3 and let it generate the output of the tool.

We observe that the final answer value is nearly always contained in the last sentence of the reasoning process. Therefore, we apply an additional GPT-3 few-shot learning process for mapping the last sentence to the answer value by prompting several sentence-answer pairs.

\section{Experiment}

% We applied our proposed framework to solve a knowledge-based numerical reasoning task.
As a proof of concept, we applied MultiTool-CoT to solve a knowledge-based numerical reasoning task.

\subsection{Dataset}
% We used the task 2 dataset of NumGLUE~\cite{mishra-etal-2022-numglue}.
% As explained in Section~\ref{sec:application}, this dataset consists mainly of numerical reasoning questions requiring chemistry knowledge.
% This dataset consists mainly of numerical reasoning questions requiring chemistry knowledge.
We used the Task 2 dataset of NumGLUE ~\cite{mishra-etal-2022-numglue}, which requires both numerical reasoning and domain-specific knowledge, mainly related to chemistry.
Example~\ref{ex:task2} shows a question in the dataset.

\ex. \label{ex:task2}
Find the amount of Calcium hydroxide required to react with 2 moles of Carbon dioxide to form 2 moles of Calcium carbonate along with 2 moles of Water.

All the answers are given as numbers.
We used 325 questions in the test split for evaluation.
We evaluated the accuracy.

% \subsection{Application to Knowledge-based Numerical Reasoning}
\subsection{External Tools}
\label{sec:application}
% As a proof of concept, we apply our proposed framework to a knowledge-based numerical reasoning task.
% We focus on the Task 2 dataset of NumGLUE~\cite{mishra-etal-2022-numglue}, which requires both numerical reasoning and domain-specific knowledge, mainly related to chemistry.
% Example~\ref{ex:task2} shows a question in the dataset.

% \ex. \label{ex:task2}
% Find the amount of Calcium hydroxide required to react with 2 moles of Carbon dioxide to form 2 moles of Calcium carbonate along with 2 moles of Water.

% In this study, we implement the following external tools and use them in the proposed framework.
We implemented the following external tools and used them in the proposed framework.

\begin{itemize}
    \item \textbf{Calculator (\textsc{Cal})}:
        The calculator is given a mathematical formula and outputs the calculation result.
        The calculator is implemented using Python's \texttt{eval} function\footnote{\url{https://docs.python.org/3/library/functions.html\#eval}}.
        Operators in mathematical formulas are replaced according to Python's syntax.
        We prompt GPT-3 to output the tool trigger, \texttt{$<<$Calculator$>>$}, with a mathematical formula on the same line.
        
    \item \textbf{Chemical reaction predictor (\textsc{Crp})}:
        The chemical reaction predictor is given the chemical formula of reactants and products and outputs the chemical reaction equation by adjusting the coefficients so that the reactants and products have the same number of each atom.
        We prompt GPT-3 to output the tool trigger, \texttt{$<<$Chemical reaction predictor$>>$}, with the reactants and products on the previous two lines.
        
    \item \textbf{Molar mass list (\textsc{Mml})}:
        The molar mass list is given a chemical formula and outputs its molar mass.
        The molar mass of the chemical formula is calculated from the atoms and their number in the formula.
        The molar mass of the atoms is obtained from the knowledge base listing the weight of all atoms.
        We prompt GPT-3 to output the tool trigger, \texttt{$<<$Molar mass list$>>$}, with a chemical formula on the same line.
\end{itemize}

\subsection{Methods for Comparison}

\begin{table}[t]
    \centering
    \begin{tabular}{lD{.}{.}{2}}
        \toprule
        Method & \\
        \midrule
        Zero-Shot$^{\dagger}$ & 1 \\
        Zero-Shot+CoT & 32.62 \\
        \midrule
        Few-Shot$^{\dagger}$  & 42 \\
        Few-Shot+CoT & 57.85 \\
        MultiTool-CoT (\textsc{Cal} only) & 62.77 \\
        MultiTool-CoT (\textsc{Crp} only) & 64.31 \\
        MultiTool-CoT (\textsc{Mml} only) & 69.23 \\
        % Few-Shot+CoT+\textsc{All} (\textbf{Ours}) & \textbf{85}.\textbf{85} \\
        MultiTool-CoT (\textbf{Ours}) & \textbf{85}.\textbf{85} \\
        \bottomrule
    \end{tabular}
    \caption{
    Performance in the Task 2 dataset of NumGLUE.
    The best result is shown in \textbf{bold}.
    ($\dagger$) is cited from \citet{mishra-etal-2022-numglue}.
    }    
    \label{tab:result}
\end{table}
\begin{figure*}[t]
    \centering
    \includegraphics[width=\textwidth]{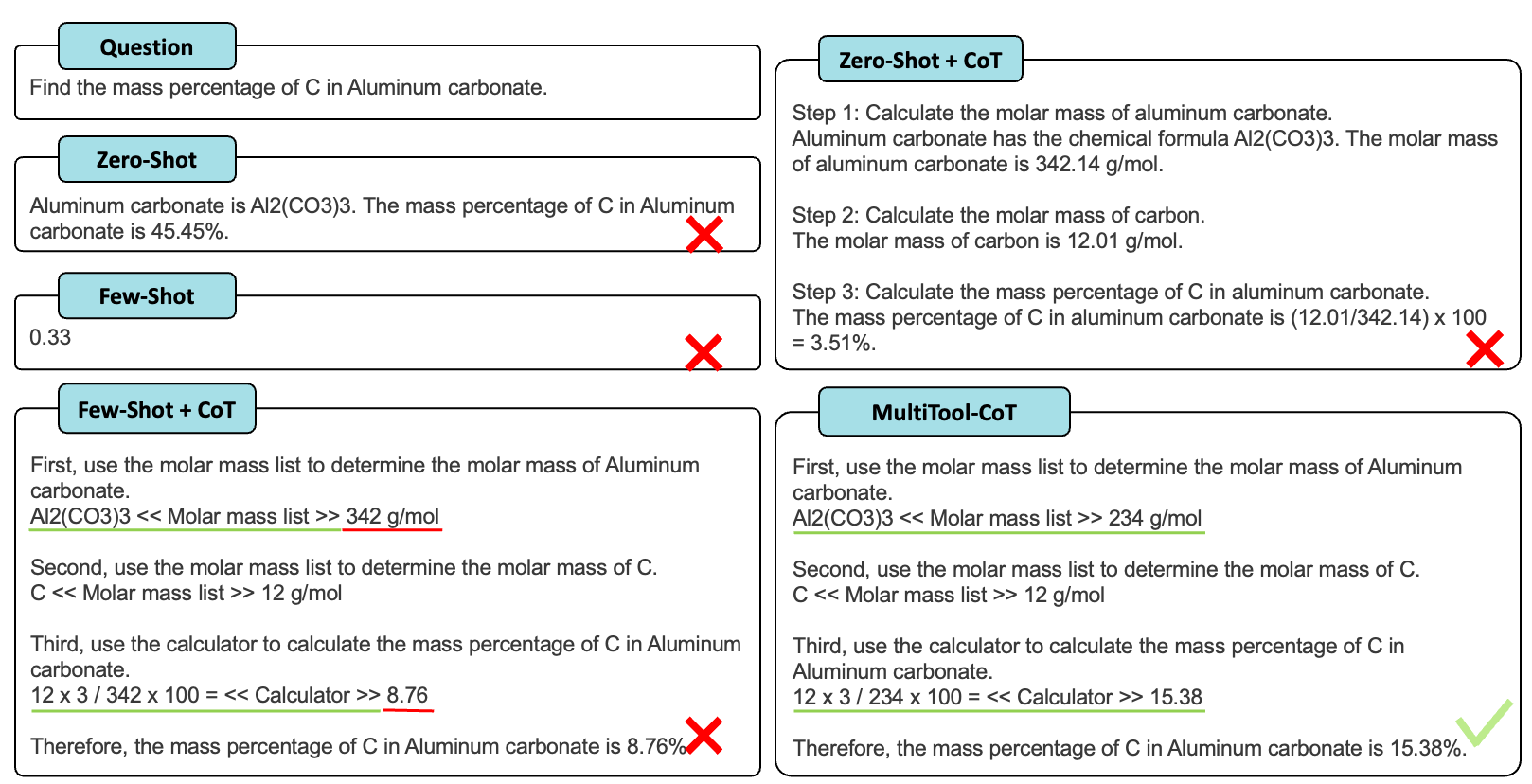}
    \caption{
    An improved example.
    % Text spans highlighted in red indicate errors related to knowledge or numerical calculation.
    % Text spans highlighted in green show improvement by using external tools.
    The green lines indicate correct reasoning processes.
    The red lines indicate errors related to knowledge or arithmetic calculation.
    }
    \label{fig:mis}
\end{figure*}

We used GPT-3 (\texttt{text-davinci-003}; 175B parameters) via OpenAI API\footnote{\url{https://openai.com/api/}} and compared the following methods.

\paragraph{Zero-Shot}
We fed only the question into GPT-3 and considered the generated text as the answer.

\paragraph{Zero-Shot+CoT~\cite{kojima2022large}}
We fed the question with the sentence ``Let's think step by step.'' into GPT-3 and obtained the answer with the intermediate reasoning steps.
We then added the sentence fragment ``Therefore, the answer (Arabic numerals) is '' after the generated text and fed it into GPT-3 to get the final answer.

\paragraph{Few-Shot}
We fed the question with few-shot examples of question-answer pairs into GPT-3 and obtained the generated text as the answer.

\paragraph{Few-Shot+CoT}
We performed the proposed method without invoking any external tools.
If the tool triggers were generated, we used GPT-3 to output the result.

\paragraph{MultiTool-CoT ($\{$\textsc{Cal}$|$\textsc{Crp}$|$\textsc{Mml}$\}$ only)}
We performed the proposed method with one of the external tools introduced in Section~\ref{sec:application}.
As for the other external tools, we let GPT-3 generate the result.

% \paragraph{Few-Shot+CoT+\textsc{All} (Ours)}
\paragraph{MultiTool-CoT (Ours)}
We performed the proposed method with all the external tools introduced in Section~\ref{sec:application}.

In few-shot settings, we used 20 questions in the training split as few-shot examples.
The questions were manually selected to avoid bias in the number of external tool calls.
% We also manually annotated the reasoning process with external tool calls.
% Zero-Shot + Let's think step by step. で出てきた推論過程を参考にし、外部ツールを利用できる箇所へ Tool trigger と その入出力を manually でアノテーションした。
% Referring to the reasoning process produced by "Zero-shot + Let's think step by step.", we also manually annotated the reasoning process with tool triggers and their inputs/outputs.
In order to annotate the questions with reasoning processes with tool triggers, we followed a two-step process.
First, we employed GPT-3 to generate the reasoning processes for solving these questions using zero-shot chain-of-thought prompting~\cite{kojima2022large}, aiming to obtain reasoning processes that GPT-3 can easily follow.
Then, we manually annotated the reasoning processes with tool triggers and the input and output for the corresponding external tools.

We set the temperature parameter of GPT-3 as 0 to generate constant predictions.
Therefore, we report the results of single runs of the methods.

\subsection{Results}

Table~\ref{tab:result} shows the results.
The proposed method achieved an accuracy of 85.85, a state-of-the-art performance.
We observed a significant performance improvement compared to methods that did not use external tools and methods that used only one external tool.
Note that the performance improvement from using multiple external tools is larger than the sum of the performance improvements from using each tool individually.
This is because GPT-3 can fail to provide accurate answers due to a combination of different types of errors, such as incorrect arithmetic calculation and knowledge.
The use of multiple external tools addressed such cases effectively, thereby improving the overall accuracy.

\subsection{Case Study}
\label{subsection:casestudy}
Figure~\ref{fig:mis} shows an improved example.
Zero-Shot and Few-Shot generated wrong answers.
Zero-Shot+CoT and Few-Shot+CoT performed reasoning based on the incorrect molar mass of Al2(CO3)3, resulting in incorrect answers.
Besides, Few-Shot+CoT failed to calculate $12 \times 3 / 342 \times 100$.
Our method, MultiTool-CoT, was able to answer correctly based on correct knowledge and calculation, relying on external tools.
More examples are presented in Figure~\ref{fig:mis3} and Figure~\ref{fig:mis4} in Appendix.

% Despite the excellent results, 14.15\% of the problems were not solved by the proposed method.
% We investigated the errors and found that they are mainly caused by incorrect or invalid tool inputs produced by GPT-3.
% Such errors mainly occurred when invoking the calculator and the chemical reaction predictor.
% Examples are shown in Figure~\ref{fig:err1} and Figure~\ref{fig:err2} in Appendix.
Despite the excellent results, there were 46 instances in which the proposed method failed to deliver accurate answers.
% 14.15\% of the examples, or 46 examples, were not solved by the proposed method.
% We manually investigated the errors and found that 18 examples were caused by incorrect reasoning processes, 16 by invalid tool inputs, 7 by incorrect NumGLUE answers, and 5 by different answer formats.
Upon manual investigation of all the errors, we identified that the majority of them were caused by incorrect reasoning processes (39\%) and invalid tool inputs (35\%).
The remaining errors were categorized into incorrect gold answers (15\%) and variations in answer formats (11\%).
% Examples of these errors are shown in Figure~\ref{fig:err1}, Figure~\ref{fig:err2}, Figure~\ref{fig:err3}, and Figure~\ref{fig:err4} in Appendix.
Examples can be found in Appendix~\ref{sec:error}.
These errors are beyond the scope of what external tools can assist with.

\section{Conclusion}
% We proposed a framework that allows LLMs to use multiple external tools, such as a knowledge retriever and a calculator, during reasoning.
We proposed MultiTool-CoT, a framework that allows LLMs to use multiple external tools, such as a knowledge retriever and a calculator, during reasoning.
% We applied the proposed framework to a numerical reasoning task that requires knowledge of chemistry and confirmed its effectiveness.
We applied MultiTool-CoT to a numerical reasoning task that requires knowledge of chemistry and confirmed its effectiveness.
The proposed framework is general and can be applied to various tasks by changing and extending external tools.
We plan to verify the effectiveness of the proposed method in other tasks in the future.

\section*{Limitations}
The major limitation of the present study is that the effectiveness of the proposed method has been confirmed only for a single task.
This is because most existing reasoning tasks are relatively simple that they can be solved by a single external tool at most.
For example, most existing numerical reasoning tasks provide self-contained questions; that is, all the required knowledge is included in the questions.
In such tasks, a calculator is all that is needed as an external tool.
However, it would be rare for a single external tool to be sufficient in real-world applications such as medical text analysis.
It is crucial for future work to validate the effectiveness in such realistic scenarios that necessitate the use of multiple external tools.
% We argue for the importance of building benchmark datasets consisting of tasks requiring the use of multiple tools to facilitate studies in this field further.

\bibliography{anthology,custom}
\bibliographystyle{acl_natbib}
\clearpage

\appendix

\section{Effect of the Number of Few-shot Examples on Performance}
\label{sec:appendix}

We investigated the effect of the number of few-shot examples on performance.
Table~\ref{tab:fewresult} shows the results.
Reducing the number of few-shot examples decreased accuracy, regardless of whether external tools were used.
Surprisingly, however, the drop in performance was not drastic, suggesting the strong generalization ability of GPT-3.
Note that it is hopeless to further improve the performance by simply increasing the number of few-shot examples because the total number of tokens in the 20 few-shot examples is nearly 3,000 while the number of tokens that GPT-3 can process is 4,000.

\begin{table}[t]
    \centering
    \begin{tabular}{lrr}
        \toprule
        & Few-Shot Examples & Acc. \\
        \midrule
        CoT & 5 & 55.38 \\
        CoT & 10 & 56.31 \\
        CoT & 20 & 57.85 \\
        \midrule
        MultiTool-CoT & 5 & 83.69 \\
        MultiTool-CoT & 10 & 84.00 \\
        MultiTool-CoT & 20 & \textbf{85.85} \\
        \bottomrule
    \end{tabular}
    \caption{
    % NumGLUE の Task 2における Few-Shot 事例の数を変えたときの性能。最良の結果を\textbf{太字}で示す。
    Performance for the different number of few-shot examples in the Task 2 dataset of NumGLUE.
    The best result is shown in \textbf{bold}.
    }
    \label{tab:fewresult}
\end{table}

% \section{Explanation of Errors}
\section{Analysis of Error Types}
\label{sec:error}
We manually investigated all 46 errors as described in Section~\ref{subsection:casestudy}. There were four types of errors: incorrect reasoning processes (39\%), invalid tool inputs (35\%), incorrect gold answers (15\%), and variations in answer formats (11\%).

\begin{figure*}[t]
    \centering
    \includegraphics[width=0.55\textwidth]{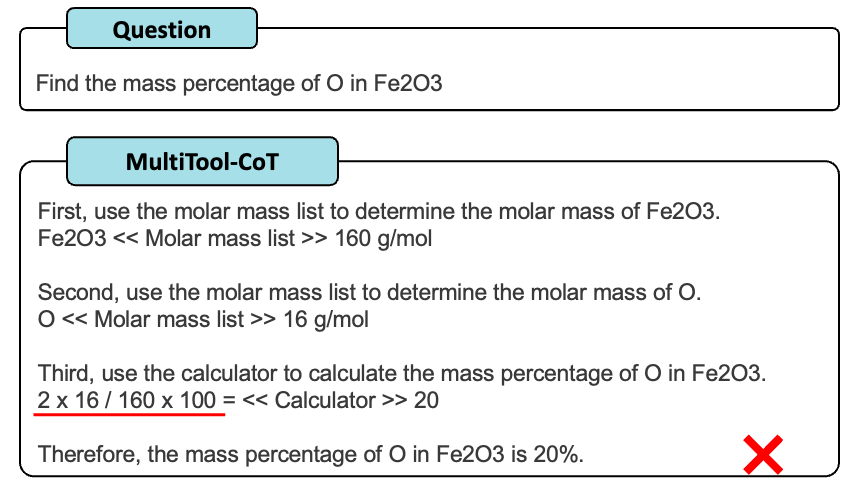}
    \caption{
        An example of incorrect reasoning processes.
    }
    \label{fig:err1}
\end{figure*}

\paragraph{Incorrect Reasoning Processes}
Figure~\ref{fig:err1} shows an error due to an incorrect reasoning process.
GPT-3 generated an incorrect mathematical formula (underlined in red), which was expected to be $3 \times 16 / 160 \times 100$.
Consequently, even though the calculation was performed correctly using the calculator, the final answer turned out to be incorrect.

\begin{figure*}[t]
    \centering
    \includegraphics[width=0.55\textwidth]{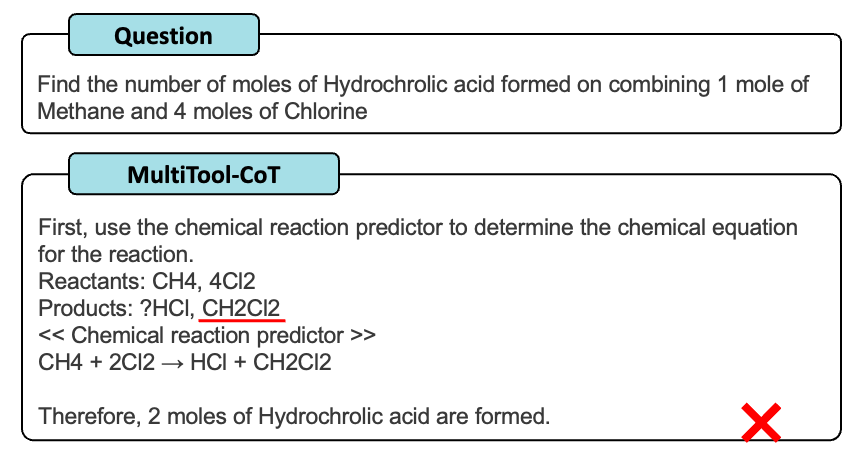}
    \caption{
        An example of the invalid tool inputs.
    }
    \label{fig:err2}
\end{figure*}

\paragraph{Invalid Tool Inputs}
Figure~\ref{fig:err2} shows an error caused by an invalid tool input.
GPT-3 generated an invalid product, CH2Cl2 (underlined in red), which was expected to be CCl4.
Thus, the chemical reaction predictor encountered a run-time error, resulting in an incorrect final answer.

\begin{figure*}[t]
    \centering
    \includegraphics[width=0.55\textwidth]{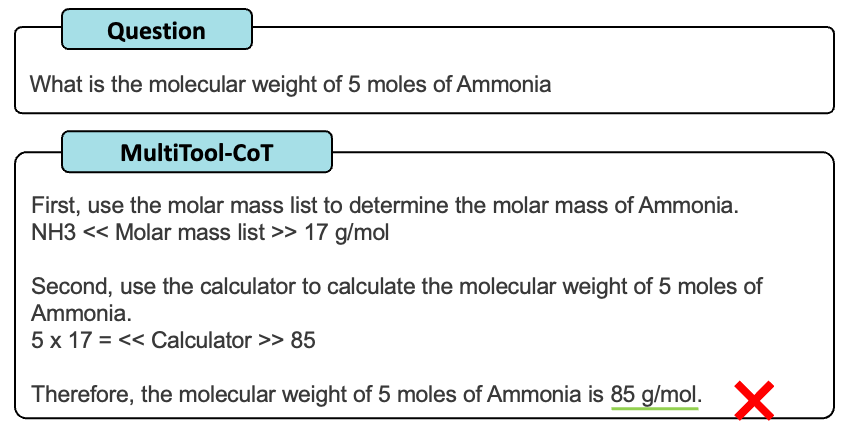}
    \caption{
        An example of incorrect gold answers.
    }
    \label{fig:err3}
\end{figure*}

\paragraph{Incorrect Gold Answers}
Figure~\ref{fig:err3} shows an error resulting from an incorrect gold answer.
The answer predicted by the proposed method was ``85 g/mol,'' whereas the gold answer was ``90 g/mol.''

\begin{figure*}[t]
    \centering
    \includegraphics[width=0.55\textwidth]{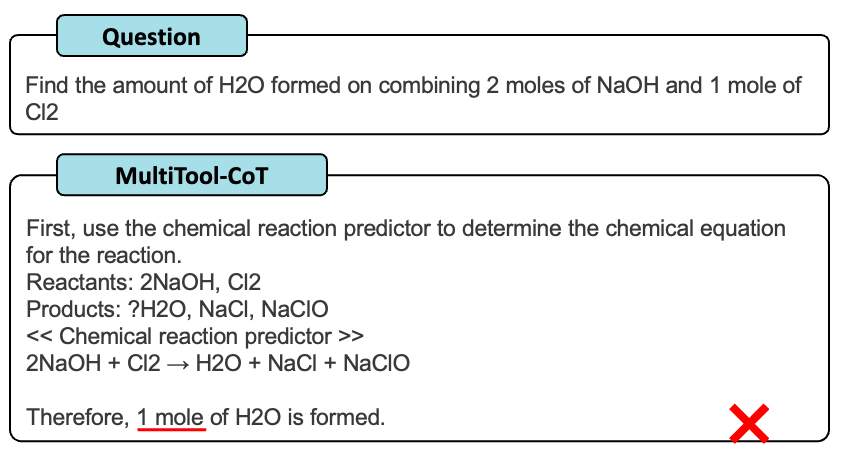}
    \caption{
        An example of variations in answer formats
    }
    \label{fig:err4}
\end{figure*}

\paragraph{Variations in Answer Formats}
Figure~\ref{fig:err4} shows an error attributed to a variation in the answer format.
The answer predicted by the proposed method was ``1 mole,'' while the gold answer was ``18 g''.
Since 1 mole of water is 18g, they both represent the same quantity.
However, due to the difference in the answer formats, it is considered an error.

\begin{figure*}[t]
    \centering
    \includegraphics[width=1.00\textwidth]{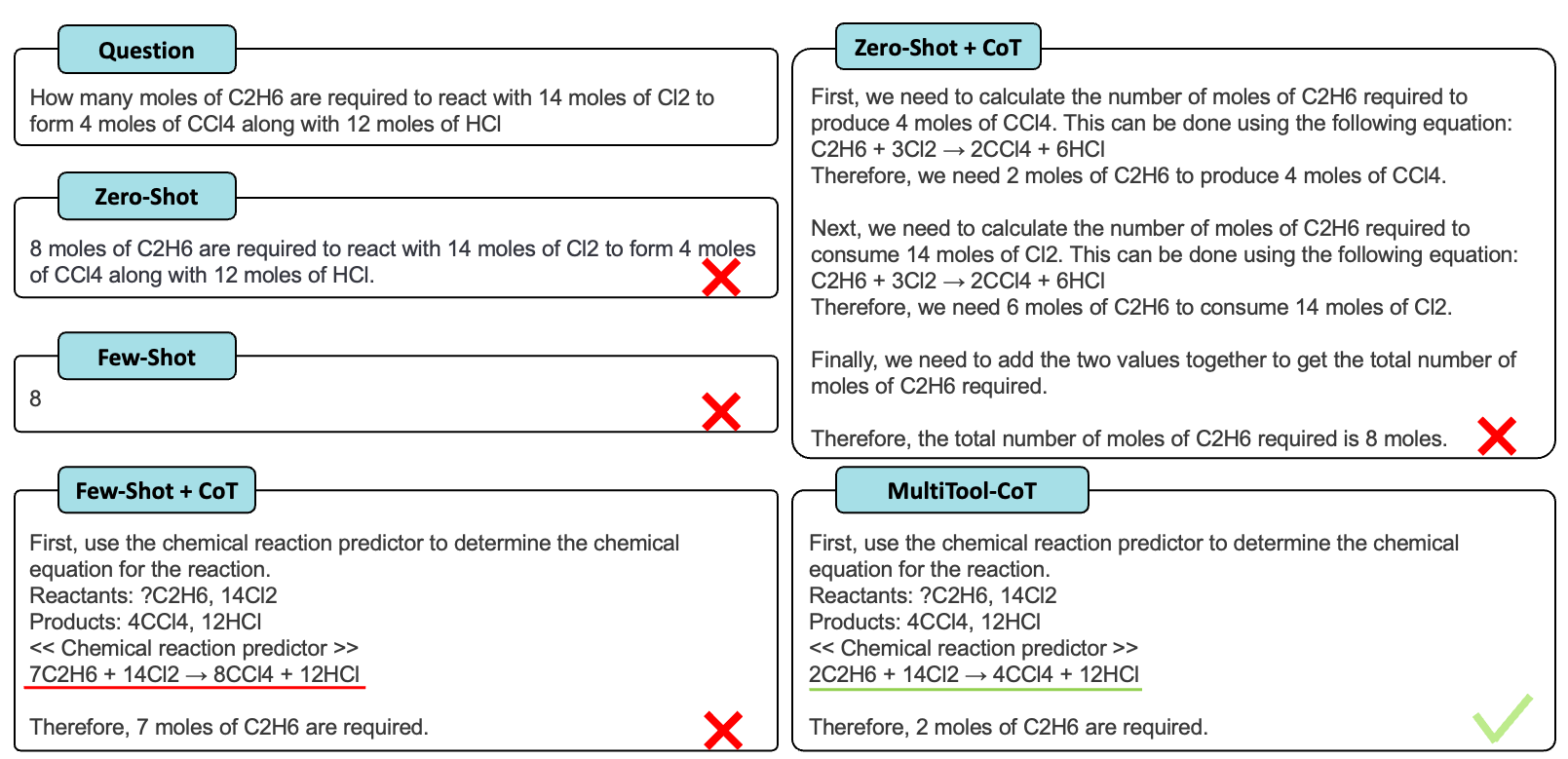}
    \caption{
        An improved example.
        % The text span highlighted in red indicates an error in chemical reaction understanding.
        % Text spans highlighted in green show improvement by using the chemical reaction predictor as an external tool.
        The red line indicates an error in chemical reaction understanding.
        The green line indicates the correct reasoning process by using the chemical reaction predictor as an external tool.
    }
    \label{fig:mis3}
\end{figure*}

\begin{figure*}[t]
    \centering
    \includegraphics[width=1.00\textwidth]{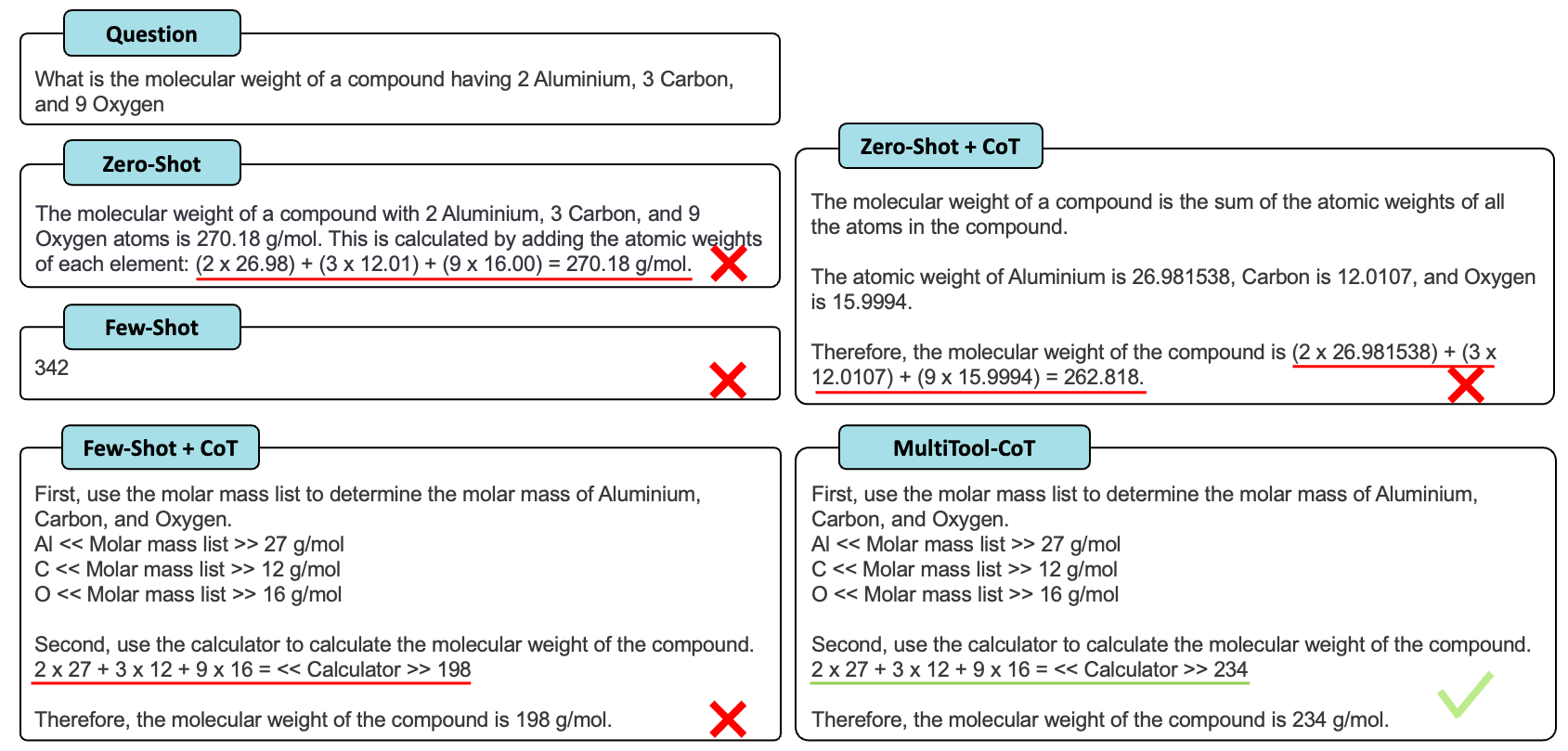}
    \caption{
        An improved example.
        % The text spans highlighted in red indicate errors in numerical calculation.
        % The text span highlighted in green shows improvement by using the calculator as an external tool.
        The red line indicates errors in arithmetic calculation.
        The green line indicates the correct reasoning process by using the calculator as an external tool.
    }
    \label{fig:mis4}
\end{figure*}

\clearpage

\end{document}